\documentclass[11pt]{article}
\usepackage{acl2015}
\usepackage{times}
\usepackage{url}
\usepackage{latexsym}
\usepackage{color}
\usepackage{amsmath}

\newcommand{\Ni}{({\em i})~}
\newcommand{\Nii}{({\em ii})~}
\newcommand{\Niii}{({\em iii})~}
\newcommand{\Niv}{({\em iv})~}

\title{Global Thread-Level Inference for Comment Classification\\in Community Question Answering}

\author{
Shafiq Joty, Alberto Barr\'on-Cede\~no, Giovanni Da San Martino, Simone Filice, 
 \\ {\bf } {\bf Llu\'is M\`arquez}, {\bf Alessandro 
Moschitti}, and {\bf Preslav Nakov}, \\
 Qatar Computing Research Institute, HBKU\\
  {\tt \{sjoty,albarron,gmartino,sfilice,}\\{\tt 
lmarquez,amoschitti,pnakov\}@qf.org.qa} 
}

\date{}

\begin{document}
\maketitle
\begin{abstract}
Community question answering, a recent evolution of question answering in the Web context, allows a user to quickly consult the opinion of a number of people on a particular topic, thus taking advantage of the wisdom of the crowd. 
Here we try to help the user by deciding automatically which answers are good and which are bad for a given question. In particular, we focus on exploiting the output structure at the thread level in order to make more consistent global decisions. More specifically, we exploit the relations between pairs of comments at any distance in the thread, which we incorporate in a graph-cut and in an ILP frameworks. 
We evaluated our approach on the benchmark dataset of SemEval-2015 Task~3. Results improved over the state of the art, 
confirming the importance of using thread level information.
\end{abstract}

\section{Introduction}
\label{sec:intro}


Community question answering (CQA) is a recent evolution of question answering, 
in the Web context,
where users pose questions and then receive answers from other users. This setup is very attractive, as the anonymity on the Web allows users to ask just about anything and then hope to get some honest answers from a number of people. On the negative side, there is no guarantee about the quality of the answers as people of very different background, knowledge, and with different motivation contribute answers to a given question.

Unlike traditional question answering (QA), 
in CQA answering takes the form of commenting in a forum. Thus, many comments are only loosely connected to the original question, and some are not answers at all, but are rather interactions between users.

As question-comment threads can get quite long,
finding good answers in a thread can be time-consuming.
This has triggered research in trying to automatically determine 
which answers might be good and which ones are likely to be bad or irrelevant. 
One early work going in this direction is that of \newcite{qu-liu:2011:IJCNLP-2011}, who tried to determine whether a question is ``solved'' or not, given its associated thread of comments. As a first step in the process, they performed a comment-level classification, considering four classes: \emph{problem}, \emph{solution}, \emph{good feedback}, and \emph{bad feedback}.

More recently, the shared task at SemEval 2015 on Answer Selection in CQA 
\cite{nakov-EtAl:2015:SemEval}, whose benchmark datasets we will use below, tackled the task of identifying \emph{good}, \emph{potential}ly useful, and \emph{bad} comments within a thread. 
%
In that task, the top participating systems used thread-level features,
in addition to the usual local features that only look at the question--answer pair.
For example, the second-best team, HITSZ-ICRC, used as a feature the position of the comment in the thread
\cite{hou-EtAl:2015:SemEval1}.
Similarly, our participation, which achieved the third-best postition, used features that try to describe a comment in the context of the entire comment thread, focusing on user interaction~\cite{nicosia-EtAl:2015:SemEval}. 
Finally, the fifth-best team, ICRC-HIT, treated the answer selection task as a sequence labeling problem and proposed recurrent convolution neural networks to recognize good comments \cite{zhou-EtAl:2015:SemEval}.

In a follow-up work, \newcite{zhou-EtAl:2015:ACL-IJCNLP} included a long-short term memory in their convolution neural network to learn the classification sequence for the thread. In parallel, in our recent work \cite{barroncedeno-EtAl:2015:ACL-IJCNLP}, we tried to exploit the dependencies between the thread comments to tackle the same task. We did it by designing features that look globally at the thread and by applying structured prediction models, such as Conditional Random Fields \cite{Lafferty01}. Our goal in this paper goes in the same direction: we are interested in exploiting the output structure at the thread level to make more consistent global assignments. 

%

To the best of our knowledge, there is no work in QA that identifies good answers based on the selection of the other answers retrieved for a question. This is mainly due to the loose dependencies between the different answer passages in standard QA. In contrast, we postulate that in a CQA setting, the answers from different users in a common thread are strongly interconnected and, thus, a joint answer selection model should be adopted to achieve higher accuracy. In particular, we focus on the relations between two comments \emph{at any distance in the thread}. This is more general than previous approaches, which were either limited to sequential interactions or considered conversational interactions only at the level of features. 


We propose a model based on the idea that similar comments should have similar labels. Below, we apply graph-cut and we compare it to an integer linear programming (ILP) formulation for decoding under global constraints; we also provide results with a linear-chain CRF.
%
We show that the CRF is ineffective due to long-distance relations, e.g., a conversation in a thread can branch and then come back later.
On the contrary, the global inference models (either graph-cut or ILP) using the similarity between pairs of comments manage to significantly improve a strong baseline performing local comment-based classifications.

\section{The Task}
\label{sec:task}

We use the CQA-QL corpus from Subtask A of SemEval-2015 Task 3 on Answer Selection in CQA. The corpus contains data from the \emph{Qatar Living} forum,\footnote{http://www.qatarliving.com/forum} and is publicly available on the task's website.\footnote{http://alt.qcri.org/semeval2015/task3/}  
The dataset consists of questions and a list of answers for each question, i.e., 
\emph{question-answer threads}.
Each question, and each answer, consist of a short title and a more detailed description. There is also meta information associated with both, e.g., ID of the user asking/answering the question, timestamp, category.
The task asks participants to determine for each answer in the thread whether it is \emph{Good}, \emph{Bad}, or \emph{Potential}ly useful for the given question.

A simplified example is shown in Figure~\ref{example},\footnote{http://www.qatarliving.com/moving-qatar/posts/can-i-obtain-driving-license-my-qid-written-employee} where answers 2 and 4 are good, answer 1 is potentially useful, and answer 3 is bad. 
In this paper, we focus on a \emph{2-class} variant of the above Subtask A, which is closer to a real CQA application. We merge \emph{Potential} and \emph{Bad} labels into \emph{Bad} and we focus on the 2-class problem: Good-vs-Bad. Table~\ref{tab:corpus} shows some statistics about the resulting dataset used for development, training and testing.

\begin{figure}[t]
{\small
\begin{description}\setlength\itemsep{0pt}
\item[Q:] Can I obtain Driving License my QID is written Employee 
\item[A$_1$] the word employee is a general term that refers to all the staff in your company either the manager, secretary up to the lowest position or whatever positions they have. you are all considered employees of your company.
\item[A$_2$] your qid should specify what is the actual profession you have. i think for me, your chances to have a drivers license is low.
\item[A$_3$] dear richard, his asking if he can obtain. means he have the driver license
\item[A$_4$] Slim chance \ldots 
\vspace*{-4mm}
\end{description}
}
\caption{\label{example}\textbf{Example from SemEval-2015 Task 3.}}
\end{figure}

\begin{table}[h]
\centering
\begin{tabular}{lrrr}
 \bf Category	& \bf Train	& \bf Dev	& \bf Test	\\\hline
  \bf Questions	& 2,600		& 300		& 329	\\
  \bf Comments	& 16,541	& 1,645		& 1,976	\\
  \,\,\,\,\,\emph{Good}	& 8,069		& 875		& 997	\\
  \,\,\,\,\,\emph{Bad}	& 8,472		& 770		& 979	\\	\hline
\end{tabular}
\caption{\textbf{Statistics about the CQA-QL dataset:} after merging \emph{Bad} and \emph{Potential} into \emph{Bad}.} 
\label{tab:corpus} 
\vspace*{-4mm}
\end{table}

\section{Our Proposed Solution}
\label{sec:method}




We
model the
pairwise relations between the comments in the answer thread ($\{c_i\}_{i=1}^n$)
to produce a better global assignment: 
we combine the predictions of a Good-vs-Bad 
classifier at the comment level with the output of a pairwise classifier, Same-vs-Different, which takes two comments and predicts whether they should have the same label. 

Each comment $c_i$ has an individual score $s_{iK}$, provided by the Good-vs-Bad classifier, for being in class $K \in \{G,B\}$ (i.e., $G$ for \emph{Good} and $B$ for \emph{Bad}). Moreover, for each pair of comments $(c_i,c_j)$, we have an association score $s_{ij}$, an estimate by the pairwise classifier about how likely it is that the comments $c_i$ and $c_j$ will have the same label. Next, we define two ways of doing global inference using these two sources of information.


\subsection{Graph Partition Approach}

Here our goal is to find a partition $P = (G,B)$ that \emph{minimizes} the following cost: 
\vspace{-1pt}
\begin{equation}
C(P) = \lambda  \Bigl[ \sum_{c_i \in G} {s_{iB}} + \sum_{c_i \in B} {s_{iG}} \Bigr] +  (1-\lambda) \hspace{-1em}\sum_{{c_i \in G},{c_j\in B}} \hspace{-1em} s_{ij} \nonumber
\end{equation}
 \vspace{-1pt}
The first part of the cost function discourages misclassification of individual comments, while the second part encourages  similar comments to be in the same class. The mixing parameter $\lambda \in [0,1]$ determines the relative strength of the two components. Our approach is inspired by \newcite{Pang:2004}, where they model the proximity relation between sentences for finding subjective sentences in product reviews, whereas we are interested in global inference based on local classifiers.

The optimization problem can be efficiently solved by finding a \emph{minimum cut} of a weighted undirected graph $G = (V,E)$. The set of nodes $V=\{v_1, v_2, \cdots, v_n, s, t\}$ represent the $n$ comments in a thread, the \emph{source} and the \emph{sink}. We connect each comment node $v_i$ to the source node $s$ by adding an edge $w(v_i, s)$ with capacity $s_{iG}$, and to the sink node $t$ by adding an edge $w(v_i, t)$ with capacity $s_{iB}$. Finally, we add edges $w(v_i,v_j)$ with capacity $s_{ij}$ to connect all pairs of comments. 

Minimizing $C(P)$ amounts to finding a partition $(S,T)$, where $S=\{s\} \cup S'$ and $T=\{t\} \cup T'$ for $s \notin S', t \notin T'$, that minimizes the cut capacity, i.e.,~the net flow crossing from $S$ to $T$. One crucial advantage of this approach is that we can use \emph{max-flow} algorithms to find the exact solution in polynomial time --- near-linear in practice \cite{Cormen:2001,Boykov:2004}.    

%

\subsection{Integer Linear Programming Approach}
Here we follow the \emph{inference with classifiers} approach by Roth and Yih~\shortcite{roth-yih:2004:CoNLL}, solved with Integer Linear Programming (ILP).
We have one ILP problem per question--answer thread. We define a set of binary variables, whose assignment will univocally define the classification of all comments in the thread. In particular, we define a pair of variables for each answer: $x_{iG}$ and $x_{iB}$, $1\leq i\leq n$. 
Assigning 1 to $x_{iG}$ means that comment $c_i$ in the thread is classified as \emph{Good}; assigning it 0 means that $c_i$ is not classified as \emph{Good}. The same applies to the other classes (here, only \emph{Bad}). Also, we have a pair of variables for each pair of comments (to capture the pairwise relations): $x_{ijS}$ and $x_{ijD}$, $1\leq i<j\leq n$. 
Assigning 1 to $x_{ijS}$ means that $c_i$ and $c_j$ have the same label; assigning 0 to $x_{ijS}$ means that $c_i$ and $c_j$ do not have the same label. The same interpretation holds for the other possible classes (in this case only \emph{Different}).\footnote{Setting a binary variable for each class label is necessary to have an objective function that is linear on the labels.}

Let $c_{iG}$ be the cost of classifying $c_i$ as \emph{Good}, $c_{ijS}$ be the cost of assigning the same labels to $c_i$ and $c_j$, etc. Following~\cite{roth-yih:2004:CoNLL}, these costs are obtained from local classifiers by taking log probabilities, i.e., $c_{iG}=-\log s_{iG}$, $c_{ijS}=-\log s_{ij}$, etc. The goal of the ILP problem is to find an assignment $A$ to all variables $x_{iG}$, $x_{iB}$, $x_{ijS}$, $x_{ijD}$ that \emph{minimizes} the cost function:
\vspace*{-6pt}
\begin{eqnarray}
C(A) &=& \lambda \cdot\sum_{i=1}^N {(c_{iG}\cdot x_{iG} + c_{iB}\cdot x_{iB})} +  \nonumber\\
     && \hspace{-3.5em} (1-\lambda) \cdot\sum_{i=1}^{N-1}  \sum_{j=i+1}^N  {(c_{ijS}\cdot x_{ijS} + c_{ijD}\cdot x_{ijD})}\nonumber
\end{eqnarray}
\vspace*{-6pt}

\noindent subject to the following constraints:
\Ni\ All variables are binary; \Nii\ One and only one label is assigned to each comment or pair of comments; \Niii\ The assignments to the comment variables and to the comment-pair variables are consistent: $x_{ijD} = x_{iG} \oplus x_{jG}, \forall i,j\;\; 1\leq i< j\leq n$.
$\lambda\in[0,1]$ is a parameter used to balance the contribution of the two sources of information.

\section{Local Classifiers}
\label{sec:classifiers}


For classification, we use Maximum Entropy, or MaxEnt, \cite{Kevin12},
as it yields a probability distribution over the class labels, which we then use directly for the graph arcs and the ILP costs.

\subsection{Good-vs-Bad Classifier}
\vspace{-2pt}
Our most important features measure the similarity between the question ($q$) and the comment ($c$). 
We compare lemmata and POS [1-4]-grams 
using Jaccard~\shortcite{Jaccard:1901}, containment~\cite{Lyon:2001}, and cosine, as well as using some similarities from DKPro~\cite{Bar2013} such as longest common substring~\cite{Allison:1986} and greedy string tiling~\cite{Wise:1996}. We also compute similarity using partial tree kernels~\cite{Moschitti:2006} on shallow syntactic trees.

Forty-three Boolean features express whether
\Ni $c$ includes URLs or emails, the words ``yes'', ``sure'', ``no'', ``neither'', ``okay'', etc., as well as `?' and `@' or starts with ``yes'' (12 features);
\Nii $c$ includes a word longer than fifteen characters (1); 
\Niii $q$ belongs to each of the forum categories (26);
and 
\Niv $c$ and $q$ were posted by the same user (4).
An extra feature captures the length of $c$.

Four features explore whether $c$ is close to a comment by the user who asked the question, $u_q$: (\emph{i-ii}) there is a comment by $u_q$ following $c$ and (not) containing an acknowledgment or (\emph{iii}) containing a question,
or (\emph{iv}) among the comments preceding $c$ there is one by $u_q$ asking a question.
%
We model dialogues 
by identifying conversation chains between two users with three features: whether $c$ is at the beginning/middle/end of a chain.
There are copies of these features for chains in which $u_q$ participates. 
Another feature for $c_{u_i}$ checks whether the user $u_i$ wrote more than one comment in the current thread. Three more features fire for the first/middle/last comment by $u_i$. One extra feature counts the total number of comments written by $u_i$. Finally, there is a feature modeling the position of $c$ in the thread.

\vspace{-2pt}
\subsection{Same-vs-Different Classifier}
\vspace{-2pt}
We use the following types of features for a pair of comments $(c_i,c_j)$: \Ni\ all the features from the Good-vs-Bad classifier (i.e., we subtracted the feature vectors representing the two comments, $|v_i - v_j|$)\footnote{Subtracting vectors is standard in preference learning~\cite{Shen:2003:SBV:1119176.1119178}. The absolute value is necessary to emphasize comment differences instead of preferences.}; \Nii\ the similarity features between the two comments, $sim(c_i,c_j)$; and \Niii\ the prediction from the Good-vs-Bad classifiers on $c_i$ and $c_j$ (i.e., the scores for $c_i$ and $c_j$, the product of the two scores, and five boolean features specifying whether any of $c_i$ and $c_j$ are predicted as \emph{Good}, \emph{Bad}, and whether their predictions are identical).


\section{Experiments and Evaluation}
\label{sec:experiments}
We performed standard pre-processing,
and we further filtered user's signatures.
All parameters (e.g., Gaussian prior for MaxEnt and the mixing $\lambda$ for the graph-cut and ILP) were tuned on the development set. We also trained a second-order linear-chain CRF to check the contribution of the sequential relations between comments.  We report results on the official SemEval test set for all methods. For the Same-vs-Different problem, we explored a variant of training with three classes, by splitting the \emph{Same} class into \emph{Same-Good} and \emph{Same-Bad}. At test time, the probabilities of these two subclasses are added to get the probability of \emph{Same} and all the algorithms are run unchanged.

\begin{table}[t]
\centering
\begin{tabular}{l@{\hspace{2mm}}cccc}
Classifier & P			& R			& F$_1$		& Acc	\\
\hline
baseline: \emph{Same}  & & & & 69.26 \\
\hline
 MaxEnt-2C	& 73.95 & 90.99 & 81.59 & 71.56 \\
 MaxEnt-3C	& 77.15 & 80.42 & 78.75 & 69.94 \\
\hline
\end{tabular}
\vspace{-.5em}
\caption{\textbf{Same-vs-Different classification.}
P, R, and F$_1$ are calculated with respect to \emph{Same}.}
\label{tab:results:same:diff}
\vspace{-1.0em}
\end{table}

Table~\ref{tab:results:same:diff} shows the results for the Same-vs-Different classification.
We can see that the two-class MaxEnt-2C classifier works better than the three-class MaxEnt-3C.
MaxEnt-3C has more balanced P and R, but loses in both F$_1$ and accuracy.
MaxEnt-2C is very skewed towards the majority class, but performs better due to the class imbalance.
Overall, 
it seems very difficult to learn with the current features, and 
both methods only outperform the majority-class baseline by a small margin.
Yet, while the overall accuracy is low, note that the graph-cut/ILP inference allows us to recover from some errors, because if nearby utterances are clustered correctly, the wrong decisions should be outvoted by correct ones.


The results for Good-vs-Bad are shown in Table~\ref{tab:results:good:bad}.
On the top are the best systems at SemEval-2015 Task 3.
We can see that our MaxEnt classifier is competitive:
it shows higher accuracy than two of them, and the highest F$_1$ overall.\footnote{This comparison is not strictly fair as the SemEval systems were trained to predict three classes, and here we remapped them to two. We just want to show that our baseline system is very strong.}

\begin{table}[h!]
\centering
\begin{tabular}{l@{\hspace{3mm}}
							cccc@{}}
System &	P			& R			& F$_1$		& Acc	\\
\hline
\multicolumn{3}{l}{\hspace*{-2mm}\bf Top-3 at SemEval-2015 Task 3}			\\
JAIST		& 80.23 & 77.73 & 78.96 & 79.10\\
HITSZ-ICRC	& 75.91 & 77.13 & 76.52 & 76.11\\
QCRI		& 74.33 & 83.05 & 78.45 & 76.97\\
\hline
\multicolumn{3}{l}{\hspace*{-2mm}\bf Instance Classifiers}			\\
MaxEnt		& 75.67	& 84.33	& 79.77	& 78.43\phantom{$\dag$}	\\
\multicolumn{3}{l}{\hspace*{-2mm}\bf Linear Chain Classifiers}			\\
CRF			& 74.89 & 83.45 & 78.94 & 77.53\\
\multicolumn{4}{l}{\hspace*{-2mm}\bf Global Inference Classifiers}	&\\
ILP				&77.04 & 83.53 & 80.15 & 79.14$\ddag$ \\
Graph-cut		&78.30 & 82.93 & \bf 80.55 & \bf 79.80$\ddag$ \\
ILP-3C			&78.07 & 80.42 & 79.23 & 78.73\phantom{$\dag$}\\
Graph-cut-3C	&78.26 & 81.32 & 79.76 & 79.19$\dag$ \\

\hline
\end{tabular}
\vspace{-.5em}
\caption{\textbf{Good-vs-Bad classification.}
$\ddag$ and $\dag$ mark statistically significant differences in accuracy compared to the baseline MaxEnt classifier with confidence levels of $99\%$ and $95\%$, respectively (randomized test).
}
\label{tab:results:good:bad}
\vspace{-1.0em}
\end{table}

The CRF model is worse than MaxEnt on all measures, 
which suggests that the sequential information does not help.
This can be because many interactions between comments are long-distance and there are gaps in the threads due to the annotation procedure at SemEval~\cite{nakov-EtAl:2015:SemEval}.

However, global inference with graph-cut and ILP improves both F$_1$ and accuracy, mostly due to better recall.
~
Graph-cut works better than ILP as it has higher precision, which helps F$_1$ and accuracy.


Both yield statistically significant improvements over the MaxEnt classifier; they also improve over the state-of-the-art JAIST system. 
Note that the devtest-tuned values of $\lambda$ for graph-cut and ILP put much lower weight to the Same-vs-Different component (values are 0.95 and 0.91, respectively).
Finally, as expected, using the predictions of MaxEnt-2C in the global classifiers is better than using those from MaxEnt-3C.

\begin{figure}[tbh]
{\small
\begin{description}\setlength\itemsep{0.5pt}

\item[Q:] I have a female friend who is leaving for a teaching job in Qatar in January. What would be a useful portable gift to give her to take with her? 

\item[A$_1$] A couple of good best-selling novels. It's hard to find much here in Doha in the way of books. 

\vspace*{-2mm}
\textbf{Local: Good, Global: Good, Human: Good} 

\item[A$_2$] ipod to entertain herself in case of boredom... a lot of patience for her students... 

\vspace*{-2mm}
\textbf{Local: Good, Global: Good, Human: Good} 

\item[A$_3$] Thanks, please keep suggestions coming, would like to send her off with a useful gift. 

\vspace*{-2mm}
\textbf{Local: Bad, Global: Bad, Human: Bad}

\item[A$_6$] Bacon. Nice bread, bacon, bacon, errmmm bacon and a pork joint.. 

\vspace*{-2mm}
\textbf{Local: Bad, Global: Good, Human: Good} 

\item[A$_9$] Couple of good novels, All time favorite movies, ..

\vspace*{-2mm}
\textbf{Local: Bad, Global: Good, Human: Good} 

\item[A$_{11}$] Ditto on the books and dvd's. Excedrin. 

\vspace*{-2mm}
\textbf{Local: Bad, Global: Bad, Human: Good} 

\item[A$_{12}$] Ditto on the bacon, pork sausage, pork chops, ham,..can you tell we miss pork! I think getting a care package together: her favorite perfume; shampoo; conditioner; glycerin soaps; set of DVDs of her favorite TV series.. Oh, and did I mention she should pack PATIENCE?

\vspace*{-2mm}
\textbf{Local: Bad, Global: Good, Human: Good}

\vspace*{-4mm}
\end{description}
}
\caption{\label{fig:discussion_example} \textbf{An excerpt of a thread} with decisions by local and global classifiers, and humans.}
\vspace{-1em}
\end{figure}

\section{Discussion}
\label{sec:ea}

We manually examined a number of examples where our global classfier could successfully recover from the errors made by the local classifier, and where it failed to do so. In Figure~\ref{fig:discussion_example}, we show the classification decisions of our local and global (graph-cut) classifiers along with the human annotations for an excerpt of a thread. 

For example, consider answers $A_6$, $A_9$, and $A_{12}$, which were initially misclassified as \emph{Bad} by the local classifier, but later recovered by the global classifier exploiting the \emph{pairwise} information. In this case, the votes received by these answers from other \emph{Good} answers in the thread for being in the \emph{same} class won against the votes received from other \emph{Bad} answers. 




Now consider $A_{11}$, which our method failed to classify correctly as \emph{Good}. Our investigation revealed that in this case the votes from the \emph{Bad} answers won against the votes from the \emph{Good} ones. The accuracy of the pairwise classifier has proven to be crucial for the performance of our overall framework. We probably need more informative features (e.g., textual entailment and semantic similarity to capture the relation between books and novels, movies and DVDs, etc.) in order to improve the pairwise classification performance.

\section{Conclusion and Future Work}
\label{sec:conclusion}

We have investigated the use of thread-level information for answer selection in CQA. We have shown that using a pairwise classifier that predicts whether two comments should get the same label, followed by a graph-cut (or ILP) global inference improves significantly over a very strong baseline as well as over the state of the art.
%
We have further shown that using a linear-chain CRF model does not help, probably because many interactions between comments are long distance. 

In future work, we would like to improve the pairwise classifiers with richer features, as this is currently the bottleneck for improving the performance in the global model. We further plan to test our framework on other CQA datasets, including on other languages.\footnote{SemEval-2015 Task 3 had an Arabic subtask, but there the answers were not coming from the same thread.}
Last but not least, we are interested in extending this research with even more global information, e.g., by modeling global decision consistency across multiple threads.

\section*{Acknowledgments}
This research was performed by the Arabic Language Technologies (ALT) group at the Qatar Computing Research Institute (QCRI), 
HBKU, 
part of Qatar Foundation.
It is part of the Interactive sYstems for Answer Search (Iyas) project,
which is developed in collaboration with MIT-CSAIL.

\bibliographystyle{acl}
\bibliography{acl15}

\end{document}